# Deep Residual Text Detection Network for Scene Text

Xiangyu Zhu, Yingying Jiang, Shuli Yang, Xiaobing Wang, Wei Li, Pei Fu, Hua Wang and Zhenbo Luo
*Machine Learning Lab*
*Samsung R&D Institute of China, Beijing*
Beijing, China
{xiangyu.zhu, yy.jiang} @samsung.com

*Abstract*—Scene text detection is a challenging problem in computer vision. In this paper, we propose a novel text detection network based on prevalent object detection frameworks. In order to obtain stronger semantic feature, we adopt ResNet as feature extraction layers and exploit multi-level feature by combining hierarchical convolutional networks. A vertical proposal mechanism is utilized to avoid proposal classification, while regression layer remains working to improve localization accuracy. Our approach evaluated on ICDAR2013 dataset achieves 0.91 F-measure, which outperforms previous state-of-the-art results in scene text detection.

## I. INTRODUCTION

Text detection is an important part of text content analysis, especially for reading natural text in the wild. Scene text detection is becoming increasing attractive to researchers, with the development of smart phone and tremendous demands of text recognition in Augmentation Reality (AR). Unlike traditional documental text, detecting scene text seems to be a much more challenging task due to illuminations, perspective distortion and complex background.

In the last few decades, series of methods [1, 2, 3] had been proposed to deal with this problem, which achieved considerable performance. Those methods can be categorized into Sliding Window based methods and Connected Component (CC) based methods. Sliding Window based method utilizes sliding windows to search the image densely for candidate text regions, and classifies text/non-text regions by traditional machine learning tools. This kind of method can be quite slow as a consequence of densely search and multi-scale windows. Comparing to the previous method, CC base method draws more attention until recently. It involved several steps, typically three. First, CCs are extracted from images as character candidates. Second, a character classifier is trained to remove the non-text CCs. Finally, remained CCs are going to be grouped into text-lines by clustering or rules. Maximally Stable Extremal Regions (MSER) is one of the most popular CC-based methods. It had been reported outstanding performance in ICDAR2013 benchmark [4]. However, the following limitations constrain its further improvement in performance. Words constituent of single character are ignored by grouping rules for the sake of precision, characters in low color contrast can not be extracted by MSER, and another disadvantage is the complex post-processing.

Convolutional Neural Networks (CNN) approach has led to a great breakthrough in object detection. Region proposal CNN (R-CNN) [5] was the first attempt to classify proposals by CNN. Then Faster R-CNN [6] was proposed, where a sub-network named RPN was designed to generate proposals autonomously by feature maps and a few additional convolution layers. Faster R-CNN used VGG-16 [7] as baseline for feature map extraction and proposal classification until deep residual network (ResNet) [8] was presented. ResNet was reported better performance in PASCAL VOC 2007 [9] and ILSVRC 2016 comparing to VGG16 and GoogLeNet [10, 11]. Moreover, the structure of ResNet was designed fully convolutional, without heavy fully connected layers. The ResNet version of Faster R-CNN was observed better performance.

Inspired by the great progress in object detection, a few CNN based methods [12, 13, 14, 15, 16, 17] had been proposed to address scene text detection. The Connectionist Text Proposal Network (CTPN) [12] is a novel framework based on Faster R-CNN, which benefits from an additional recurrent neural network and vertical proposal mechanism.

In this paper, we came up with a framework called Residual Text detection Network (RTN). RTN were inspired by ResNet and CTPN vertical proposal mechanism. First, ResNet was used to generate strong semantic feature instead of traditional networks like VGG-16. Rather than a naively layer replacement, we combine multi-level features to produce hierarchy residual feature. The outstanding performance was mainly contributed by this stronger semantic feature. Second, vertical proposal mechanism was adopted and an additional regression part was used to improve localization accuracy, this step was implemented by a two stage training strategy. It achieved 91.54% F-measure on ICDAR2013.

## II. RELATED WORK

### A. Object detection

With the success of deep convolutional network in image recognition, R-CNN was inspired to classify region proposal via CNN. After R-CNN was proposed, the related object detection approaches had been developed rapidly, such as SPP-net, Fast R-CNN, Faster R-CNN, R-FCN. Faster R-CNN is a mature prevalent framework that trained and tested from end to end. The framework constitutes of three parts: (1) Feature map generation. Feature maps representing semantic information were extracted by deep convolutional network, VGG-16 was used in Faster R-CNN. (2) Proposal generation. A simple convolutional network name Region Proposal Networks (RPN) was designed to generate candidate regions with the input of feature maps. (3) Region classification and regression. By sharing features regions proposals were projected to the location in feature maps, then a following Fast R-CNN structure outputted final results by classification and regression.

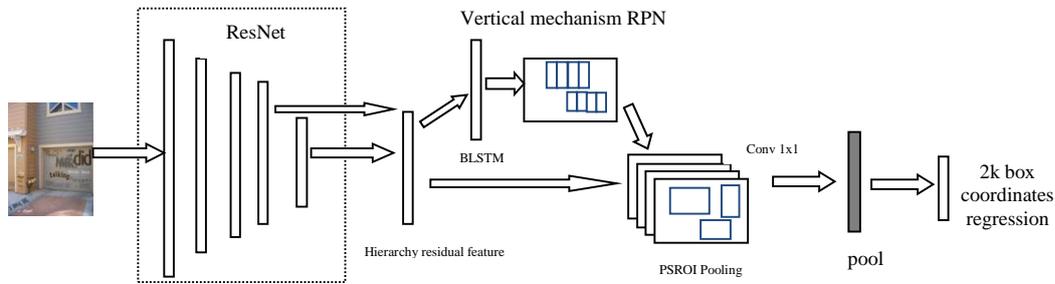

Fig.1. Architecture of Residual Text detection Network (RTN)

Influenced by the latest progress in image recognition, other deeper convolutional networks were transplant to this framework instead of VGG-16 [7], including GoogLeNet [10, 11] and ResNet [8]. ResNet was proved to be a superior convolutional network than GoogLeNet and VGG16 in ImageNet classification task. R-FCN [18] is a completely fully convolutional architecture that combines ResNet and Faster R-CNN together. FPN (Feature Pyramid Network) [19] exploits multi-scale pyramid of ResNet, and the framework using FPN won the champion of COCO [20] detection challenge 2016.

Besides Faster R-CNN based pipeline, Single Shot MultiBox Detector (SSD) [21] and You Look Only Once (YOLO) [22] are two representative and promising works. SSD is one of the first attempts to utilizing multi-level convolutional networks, while YOLO is extremely faster than all the methods mentioned above. However, they do not get a superior performance with significant margin comparing to Faster R-CNN pipeline.

### B. CNN based text detection

General object detection pipeline can be transplant to text detection realm barrier free. CNN based text detection gradually becomes the most promising approach. Zhang [14] proposed a fully convolutional network for text detection in arbitrary orientation instead of semantic segmentation. It achieved an F-measure of 0.74 on ICDAR2013. DeepText [13] proposed a Inception-RPN and multi-level region-of-interest pooling based on the framework of Faster R-CNN. It achieved 0.85 F-measure on ICDAR2013. Inspired by SSD, Liao [15] presented a approach called TextBoxes, multi-level jointly predictions and word recognition were utilized.

CTPN [12] is a unique network abandoned Fast R-CNN classification and regression, which can be treated as a novel individual RPN with Recurrent Neural Network (RNN). It achieved previous state-of-the-art on ICDAR2013 as 0.88 F-measure among published papers. Nevertheless, it was just a prototype for detection using RNN and fixed width proposal is harmful for localization accuracy.

## III. RESIDUAL TEXT DETECTION NETWORK

The architecture of this Residual Text detection Network (RTN) is shown in Fig. 1. It consists of three parts: hierarchy residual feature map for feature extraction, vertical mechanism RPN for proposal prediction and bounding box regression part for higherlocalization accuracy.

### A. Hierarchy residual feature map

In our framework, we use ResNet to derive feature map from original images. The feature map is a serial of features in 2D formation, similar to handcraft feature. It is fed to RPN and regression part. ResNet consists of 5 concatenate blocks (i.e., conv1, conv2_x, conv3_x, conv4_x and conv5_x). Conv4_x have the same stride as VGG-16 output (16 pixels). In R-FCN[18], region proposals were predicted by conv4_x. They believed the conv4_x feature maps were semantic strong enough and comparable to VGG-16 feature maps. VGG-16 differs from ResNet in structure. Thus, a simple replacement, from VGG-16 to ResNet, would not work properly. Unlike VGG-16, typical ResNet based detection does not share the same feature map between RPN and regression parts. Conv4_x is utilized to generate proposals in RPN while conv5_x for regression. In this kind of methods, RPN is unable to use a deeper semantic feature. By visualizing feature maps of conv3_x, conv4_x, conv5_x and VGG-16, we find out conv3_x contains too many low level features, while conv4_x and conv5_x are competitive to VGG-16 on the first glance. We have carried out series of experiments on Faster R-CNN baseline using conv3_x, conv4_x and conv5_x respectively. Framework using conv3_x detected edges and lines instead of objects and required much more computation due to larger feature map sizes. It was a strong evidence that conv3_x contained too many low level features to be used directly. On the contrary, baselines using conv4_x and conv5_x detected text correctly. However, framework using conv5_x fails on detecting small text due to coarse resolution feature maps. Although conv5_x represents deeper feature, the resolution is half comparing to conv4_x. Even we adopt the "*à trous algorithm*" [23] to compensate stride difference, the performance is still unsatisfactory. Using conv5_x as the only feature maps might be insufficient for text detection, but abandon deeper representations seems to be an unwise choice. We believe using conv5_x in a proper way will contribute to proposal prediction.

It is rational to come up with a naive idea that predicting multi-scale proposals on conv4_x and conv5_x respectively,

like previous approaches did, such as SSD and TextBoxes. In this way, not only we can detect fine scale text and robust to scale invariance, but also utilizing deeper feature representations. Nevertheless, it is inconvenient to identify reliability from multi-scale proposals without an additional classification, as we introduced vertical mechanism to RPN, it seems to be a rather complicated problem.

To deal with that, we combine the hierarchy feature maps (conv4_x and conv5_x) together to produce a new hierarchy feature map. In this way, we can use both conv4_x and con5_x feature maps simultaneously, and the task to identify which feature maps are more reliable is assigned to convolution layers. As shown in Fig 2, the input size of original images is $224 \times 224$, after several convolutional layers, conv4_x and conv5_x get feature maps in size $14 \times 14$ and $7 \times 7$, corresponding to 16 pixels and 32 pixels stride. A deconvolution layer was used to upsample conv5_x, make sure the shapes of conv5_x (res5c) match conv4_x (res4b22) exactly. We attach a convolution layer with kernel $1 \times 1$, which aim to work as learnable weights for combining conv5_x and conv4_x. Our experiment shows hierarchy feature lead to an improvement on both precision and recall.

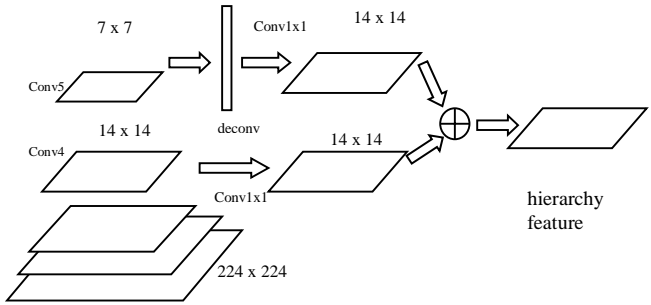

Fig.2. Hierarchy residual network architecture. First, conv5_x was upsampled to make sure its shapes match conv4_x. Second, we attached convlutional layers with kernal $1 \times 1$ to both conv5_x and conv4_x. Finally, hierarchy feature was produced by element wise addition.

### B. Vertical mechanism RPN

In Faster R-CNN, a serial of CNN is used to classify proposals. The structure is called Fast R-CNN [24]. However, CTPN abandoned Fast R-CNN structure, namely RPN output vertical proposals directly without classification and regression. As we know, RPN can be treated as a general object detection system. If the detection task is to distinguish only one category from background (two categories in total), it seems that RPN is already competent for text detection. Depending on vertical proposal mechanism and recurrent neural network, CTPN [12] was able to detect text without Fast-RCNN. That mechanism makes the final model much smaller.

In this approach, we adopt this vertical mechanism to RPN. Anchors and ground truth are divided into fixed width (16 pixels) boxes, shown in Fig 3. Particularly, spaces between ground truths are treated as negative samples. This enable the method to output result in word level. Sequences of vertical proposals will be predicted by RPN. A threshold is applied to remove non-text vertical proposals, therefore remained adjacent text proposals can be connected together to produce text line proposals.

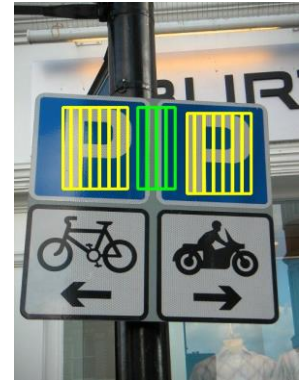

Fig.3 Yellow box: ground truth of vertical proposals. Green box: space between words which are treated as negative samples

### C. Bounding box regression

By connecting vertical proposals, we will obtain text-line proposals as result. Nevertheless, fixed width proposal might lead to inaccurate localization, when the beginning and the end of vertical proposals are not exactly fit text. In small text case, the problem becomes more serious. Unlike general object detection, this inaccuracy will influence recognition tremendously. If parts of the characters are not included in bounding box, they might be omitted or wrongly recognized. On the contrary, a loose bounding box contains much background, and that could be recognized as additional characters. In conclusion, a tight and exact bounding box is significant for text detection and recognition.

To achieve this goal, we introduce bounding box regression to get exact coordinates, just as Faster R-CNN and R-FCN did in their framework. In this paper, we refer to Fast R-CNN structure. As text line proposals are obtained in section B, bounding box offset of every proposal were calculated. However, classification is not contained in this part, only regression is remained. A further classification is unnecessary, and experiments show it is harmful for performance. This is because recurrent neural networks we adopted in RPN have a tendency to connect words into text lines. After we set word level as network learning goal, text line level proposal might be classified as negative result.

The bounding box regression loss is defined as:

$$L_{loc}(t^u, v) = \sum_{i \in \{x,w\}} smooth_{L_1}(t_i^u - v_i)$$

$$smooth_{L_1}(x) = \begin{cases} 0.5x^2 & if |x| < 1 \\ |x| - 0.5 & otherwise \end{cases}$$

In this functions, $v = (v_x, v_w,)$ is the ground truth of bounding box, $t^u = (t_x, t_w,)$ is predicted coordinates, $x$ and $w$ stand for x coordinate and width. $L_1$ smooth function is used for regression. This loss function is almost the same as what used in Fast R-CNN except that two coordinates $(x, w)$ offsets are predicted instead of four coordinates $(x, y, h, w)$. It is unnecessary to regression y coordinate $(y)$ and height $(h)$ which was done in RPN layers for every single vertical proposal.

We develop a two stage training strategy to implement this further regression:

- Stage one. Hierarchy residual feature and vertical mechanism RPN were trained; the learning rates of regression parts were set to 0.
- Stage two. Regression parts were trained individually, the learning rates of ResNet, hierarchy residual feature and RPN were set to 0. A normal RPN presented in Faster R-CNN is used to generate anchors and train regression parts and will not be used in test model.

*D. Training and testing details*

Our model was trained on 15,000 natural images collected and labeled by ourselves. These images were labeled in word level and resized to (600, 1000) scale. There is no overlap between these images or any kind of public dataset available on the internet. On the condition of the extremely similarity between ICDAR2013 training set and testing set, ICDAR2013 training set was not included to prevent over-fitting ICDAR2013 testing set.

Training ground truth is labeled in word level, and then divided into vertical ground truth by a fixed width (16 pixels) in proposal layer, corresponding to vertical proposals mentioned above. The space between words were labeled as negative samples, anchor has an IoU> 0.5 overlap with space samples were signed as negative label. About 10% of negative samples are space. By adding space sample, the networks tend to output word level proposal rather than text-line level.

## IV. EXPERIMENTS

We evaluated RTN on ICDAR 2013 benchmarks. It consists of 233 focused text images taken in the wild. The evaluation criteria are provided by the ICDAR2015 Robust Reading Competition website[1] as previous works did.

First, the effectiveness of hierarchy residual feature map was verified comparing to other prevalent feature extraction layers. Then, additional regression layers were proved to be helpful for localization accuracy. Finally, this method was compared to other published methods, and it achieved state-of-the-art performance.

*A. Evaluation of hierarchy residual feature*

In CTPN, 3,000 natural images were collected and label for training, much less than ours. In order to prove the improvement is a consequence of stronger semantic feature map rather than much more training data, we implement our own version of CTPN and training on 15,000 images. All the experiments carried out below were trained on the same amount of images.

In this experiment, we used VGG-16 and ResNet-101 as backbone for feature extraction. Feature map generated by different layers were evaluated, including conv5 of VGG-16, conv4_x (res4b22) of ResNet-101 and hierarchy residual feature map (res4b22 + res5c) used in RTN.

Table 1 shows the performances on ICDAR2013, we use CTPN framework as baseline and different feature maps mentioned above are evaluated, all the parameters and following processing are the same. We evaluated these methods on two scales respectively, namely (600, 1000) and (960, 1280). Scale (600, 1000) means the shortest side of images is no more than 600 pixels and the longest side can not exceed 1000 pixels, so does to scale (960, 1280).

One observation is that all these feature maps are competitive in scale (600, 1000). However, when it comes to scale (960, 1280), margins between these methods becoming considerable. We had run the open source test code[2] provided by the author of CTPN, marked as CTPN-tianzhi. Larger scale did not benefit performance. On the contrary, the F-score degraded. Moreover, our CTPN implementation with VGG-16 improved slightly on F-scores. In conclusion, larger test scale does not always helpful for detection and localization. Nevertheless, by simply replacing VGG-16 to ResNet-101 conv4_x (res4b22), F-score improved to from 88.75% to 90.32%, it proves ResNet-101 has a superior feature representation comparing to VGG-16 as other papers [8, 18] mentioned. Furthermore, baseline with hierarchy residual feature map (res4b22 + res5c) achieved the best performance with F-score=91.17%, which improve 5 points on recall comparing to the original CTPN.

The results shows baseline with hierarchy residual feature achieves the best performance on both recall and precision on scale (960, 1280), which could be a convincing evidence for stronger semantic feature.

TABLE 1. EVALUATING BASELINE WITH DIFFERENT FEATURE MAP ON ICDAR2013

| Method | Backbone | Scale | Feature map | Precision | Recall | F score |
| --- | --- | --- | --- | --- | --- | --- |
| CTPN-Tianzhi | VGG-16 | (600,1000) | conv5 | 92.98% | 82.98% | 87.69% |
| CTPN-Tianzhi | VGG-16 | (960,1280) | conv5 | 91.02% | 82.98% | 86.81% |
| CTPN | VGG-16 | (600,1000) | conv5 | 92.56% | 83.96% | 88.06% |
| CTPN | VGG-16 | (960,1280) | conv5 | 91.52% | 86.14% | 88.75% |
| CTPN | ResNet-101 | (600,1000) | res4b22 | 93.38% | 82.76% | 87.75% |
| CTPN | ResNet-101 | (960,1280) | res4b22 | 93.62% | 88.09% | 90.32% |
| RTN | ResNet-101 | (600,1000) | res4b22+res5c | **93.65%** | **83.14%** | **88.08%** |
| RTN | ResNet-101 | (960,1280) | res4b22+res5c | **93.64%** | **88.82%** | **91.17%** |

- 

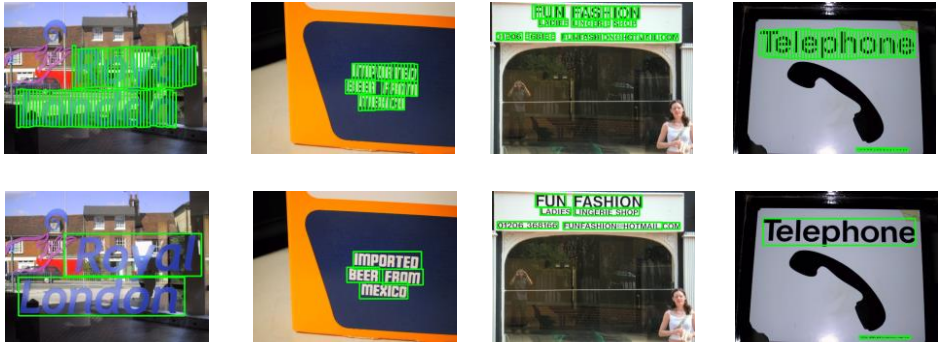

Fig.4. Example detection results of our RTN on the ICDAR2013 benchmark. The first row of images is the result before connection and regression. The second row of images is the result after vertical proposal connection and regression.

TABLE 2 REGRESSION IMPROVEMENT BY ADDITIONAL REGRESSION

| Convolutional layers | Precision | Recall | F score |
|---|---|---|---|
| RTN_no_regression | 93.64% | 88.82% | 91.17% |
| RTN_regression | **94.20%** | **89.02%** | **91.54%** |

### B. Regression improvement

Proposals connected by fixed width vertical proposals are inaccurate on both beginning and end sides. Moreover, the evaluation criteria are extremely strict. Detection bounding box can be judged as false positive sample if its boundary exceed ground truth slightly. It means this inaccuracy can degrade performance on both recall and precession, even if the texts are detected correctly.

Through bounding box regression, we are able to deal with this problem properly. As shown in Table 2, RTN with regression improved 0.4% on F-scores, both recall and precision benefit from this additional regression.

### C. Evaluation of RTN on ICDAR2013

After proving the effectiveness of hierarchy residual feature and additional regression, we compare RTN with other published methods on ICDAR2013. This single model approach did not utilize multi-scale training and multi-scale testing. Running time of each image is about 0.8s with GPU. Fig4 shows examples of detection results on ICDAR2013.

First, we compared RTN with methods mentioned in recent publications. CNN based text detection methods were compared, including Textboxes, DeepText, Multi-oriented FCN, CCTN, SegLink and CTPN. The prevalent object detection frameworks like Faster R-CNN and R-FCN are also evaluated. Table 3 shows RTN achieved the best performance with great margin.

Second, we submitted our results to ICDAR2015 Robust Reading Competition website and compared RTN with other competitors on CHALLENGE 2.1. This task is also evaluated on ICDAR2013 dataset. RTN with single model ranked third performance with slightly margin (F-scores=0.3%) compared to "Tencent Youtu" and "NLPR-CASIA".

TABLE 3 COMPARISON WITH STATE-OF-THE-ART PUBLICATIONS ON ICDAR2013

| Method | Precision | Recall | F score |
|---|---|---|---|
| Yin [1] | 0.88 | 0.66 | 0.76 |
| Faster R-CNN baseline[6] | 0.86 | 0.75 | 0.80 |
| R-FCN[18] | 0.90 | 0.76 | 0.83 |
| Multi-Oriented-FCN[14] | 0.88 | 0.78 | 0.83 |
| SegLink[17] | 0.87 | 0.83 | 0.85 |
| DeepText[13] | 0.87 | 0.83 | 0.85 |
| TextBoxes[15] | 0.89 | 0.83 | 0.86 |
| CCTN[16] | 0.90 | 0.83 | 0.86 |
| CTPN[12] | 0.93 | 0.83 | 0.88 |
| **Proposed RTN** | **0.94** | **0.89** | **0.91** |

## V. CONCLUSIONS

In this paper, a deep residual text detection network is proposed based on the prevalent object detection framework. First, stronger semantic feature is obtained by using deep residual networks and combining multi-level feature from different convolutional networks. Then, a vertical proposal mechanism is introduced inRPN inspired by CTPN. At last, an additional regression system is used to improve localization accuracy.

TABLE 4 COMPARISON WITH STATE-OF–THE-ART SUBMISSIONS ON COMPETITION WEBSITES

| Method | Precision | Recall | F score |
|---|---|---|---|
| Tencent Youtu | 94.26 % | 89.53 % | 91.84 % |
| NLPR-CASIA | 94.63 % | 89.17 % | 91.82 % |
| **RTN** | **94.20 %** | **89.02 %** | **91.54 %** |
| RRPN-4 | 95.19% | 87.31 % | 91.08% |
| MSRA_v1 | 93.67% | 88.58% | 91.06% |
| Baidu IDL | 92.83% | 87.11% | 89.88% |